\documentclass{article}
\usepackage{fullpage}

\usepackage[utf8]{inputenc} 
\usepackage[T1]{fontenc}    
\usepackage{url}            
\usepackage{booktabs}       
\usepackage{amsfonts}       
\usepackage{nicefrac}       
\usepackage{microtype}      
\usepackage{soul}
\usepackage{graphicx}
\usepackage{amsmath}
\usepackage{outlines}
\usepackage{xurl}

\usepackage[colorlinks=true,
citecolor=blue,
filecolor=black,
linkcolor=blue,
urlcolor=blue]{hyperref}

\usepackage{amsmath,amsfonts,bm}

\usepackage{xcolor}
\usepackage{colortbl}

\def\eqref#1{equation~\ref{#1}}

\DeclareMathAlphabet{\mathsfit}{\encodingdefault}{\sfdefault}{m}{sl}
\SetMathAlphabet{\mathsfit}{bold}{\encodingdefault}{\sfdefault}{bx}{n}

\usepackage{multirow}
\usepackage{paralist}

\usepackage{multicol}
\usepackage{diagbox}

\usepackage[ruled,noend]{algorithm2e}

\SetCommentSty{mycommfont}

\usepackage{here}

\usepackage{amsmath,amssymb,amsfonts,amsbsy,amsfonts,latexsym}
\usepackage{makecell}
\usepackage{xcolor}
\usepackage{colortbl}

\usepackage{tabularx,colortbl,xcolor}
\usepackage[normalem]{ulem}
\useunder{\uline}{\ul}{}

\usepackage{enumitem}

\usepackage{xparse}

\SetKwInput{KwInput}{Input}
\SetKwInput{KwRequire}{Require}

\usepackage{longtable}

\NewDocumentCommand{\var}{O{s} m O{}}{%
  \ensuremath{#1_{#2}^{#3}}
}
\usepackage{siunitx}

\newcommand{\commentout}[1]{}

\definecolor{light-gray}{gray}{0.80}

\newcommand\tref{Table~\ref}

\usepackage{amsthm}

\usepackage{amsthm}

\newtheorem{mytheorem}{Theorem}
\newtheorem{assumption}[mytheorem]{Assumption}
\newtheorem{lemma}[mytheorem]{Lemma}
\newtheorem{remk}[mytheorem]{Remark}

\usepackage{subfig}

\graphicspath{ {./figures/} }

\usepackage{pifont}

\usepackage{chngcntr}
\usepackage{adjustbox}
\usepackage{longtable}

\begin{document}


\title{
Selective Guidance: Are All the Denoising Steps of Guided Diffusion Important?
}

\author{
Pareesa Ameneh Golnari, Zhewei Yao, and Yuxiong He
\\  Microsoft \\ 
{\tt \small\{pagolnar, zheweiyao, yuxhe\}@microsoft.com}
}

\date{}
\maketitle


\begin{abstract}
This study examines the impact of optimizing the Stable Diffusion (SD) guided inference pipeline. We propose optimizing certain denoising steps by limiting the noise computation to conditional noise and eliminating unconditional noise computation, thereby reducing the complexity of the target iterations by 50\%. Additionally, we demonstrate that later iterations of the SD are less sensitive to optimization, making them ideal candidates for applying the suggested optimization. Our experiments show that optimizing the last 20\% of the denoising loop iterations results in an 8.2\% reduction in inference time with almost no perceivable changes to the human eye. Furthermore, we found that by extending the optimization to 50\% of the last iterations, we can reduce inference time by approximately 20.3\%, while still generating visually pleasing images.

\end{abstract}
\section{Introduction}
\label{sec:intro}
Diffusion models are a cutting-edge class of generative models that are designed to generate high-resolution images with a diverse range of features. Several example architectures have been proposed for diffusion models, including GLIDE~\cite{GLIDE}, DALLE~\cite{DALLE}, and Imagen~\cite{imagen}. Additionally, the full open-source Stable Diffusion (SD) architecture~\cite{SD} has gained significant attention in recent years due to its impressive performance. In this note, we will focus specifically on the SD architecture and its implementation using the Huggingface pipeline~\cite{HF-SD}.

\subsection{Guided diffusion}
Guided image generation requires a critical consideration of how the sampling process can be conditioned to a particular condition such as a text embedding. This process is commonly referred to as \textit{guided diffusion} and is essential in a variety of applications, such as text-to-image generation, image-to-image generation, and in-painting. In some models, such as SD, this process can be performed using classifier-free guidance~\cite{classifierfree} where the noise is computed in the following manner:
\begin{equation}
    \label{guided-diffusion}
        \hat{\epsilon_\theta}(x_t|y)=\epsilon_\theta(x_t|0)+s.(\epsilon_\theta(x_t|y)-\epsilon_\theta(x_t|0)).
\end{equation}
Here \textit{s} scales the perturbation, $\epsilon_\theta(x_t|0)$ is the unconditional term and $\epsilon_\theta(x_t|y)$ is the conditional term. For a more detailed explanation of the mathematical concepts behind this formula refer to~\cite{classifierfree} and~\cite{the-ai-summer}.

\subsection{Proposed optimization}
We propose a novel approach for optimizing guided image generation by simplifying the noise computation process. In applications that the guidance scale is larger than one, the conditional noise has a more significant impact than the unconditional noise in the diffusion process. Therefore, we suggest eliminating the unconditional noise in some denoising iterations to cut the computational complexity of that iteration in half.

Furthermore, we recommend limiting the optimization to the less sensitive iterations to avoid any adverse effects on the overall performance of the model. Our analysis indicates that the later iterations of SD pipeline are ideal for this optimization.
We have applied this optimization to the SD example in the DeepSpeedExamples \cite{DeepSpeedExamples-repo}, and the results demonstrate a considerable improvement in computational efficiency without sacrificing the overall quality of the generated images. This is further discussed in section \ref{sec:experiments}.
\section{Sensitivity of iterations}
\label{sec: sensitivity}
In order to produce a single image, the denoising loop is executed multiple times, usually ranging from 50 to 200 iterations for SD. Each iteration serves a distinct purpose in the process; while the initial iterations establish the general layout of the image, the later ones are dedicated to refining its overall quality. Consequently, we propose that the later iterations are comparatively less sensitive to optimization than the early ones. This claim is substantiated in this section.

\begin{figure}
  \centering
  \includegraphics[width=1.0\textwidth]{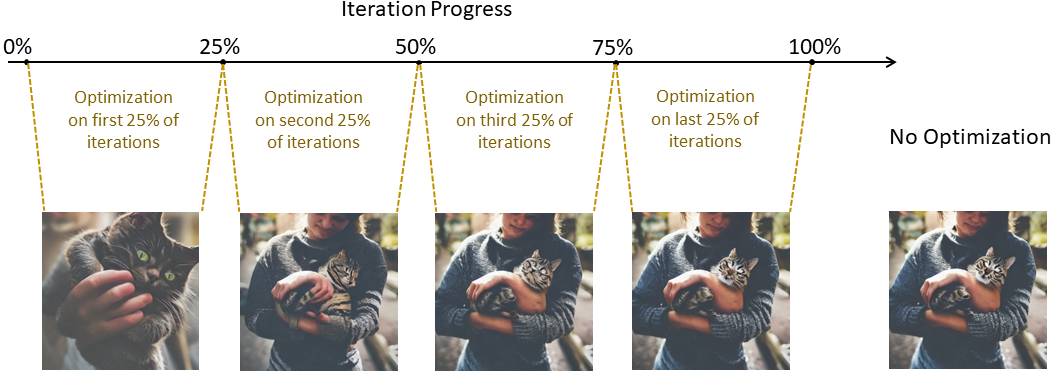}
  \caption{Images generated for the prompt “A person holding a cat”. The leftmost image was generated using optimization solely on the first 25\% of the total iterations. For the subsequent images, the optimization window gradually shifts towards the right.}
  \label{sensitivity-fig}
\end{figure}

Figure \ref{sensitivity-fig} showcases four images that were generated using the same prompt, “\textit{A person holding a cat}”. These images were created using identical parameters, with the exception of the location of the optimization window.
Upon examination of the four images, it is apparent that the quality of the generated image increases as the optimization window shifts to the right. This observation provides further confirmation of the decreased sensitivity of later iterations to optimization. It should be noted that the computational benefit is uniform across all four experiments, as the same number of iterations are optimized in each case. As a result, optimizing later iterations while preserving earlier ones is a viable strategy, and one that we will adopt for the remainder of this study.
\section{Experiments}
\label{sec:experiments}
To explore the effects of optimization, we created a series of images with various prompts. These images were generated using different degrees of optimization to measure its impact on the final output while keeping all other variables constant. Denoising iterations were fixed at 50, and the random seed was held constant. When conducting the optimization experiments, we directed our efforts towards optimizing the final iterations, given our earlier discovery that these iterations exhibited the least sensitivity to denoising.

\subsection{The optimization effect}
\begin{figure}
  \centering
  \includegraphics[width=1.0\textwidth]{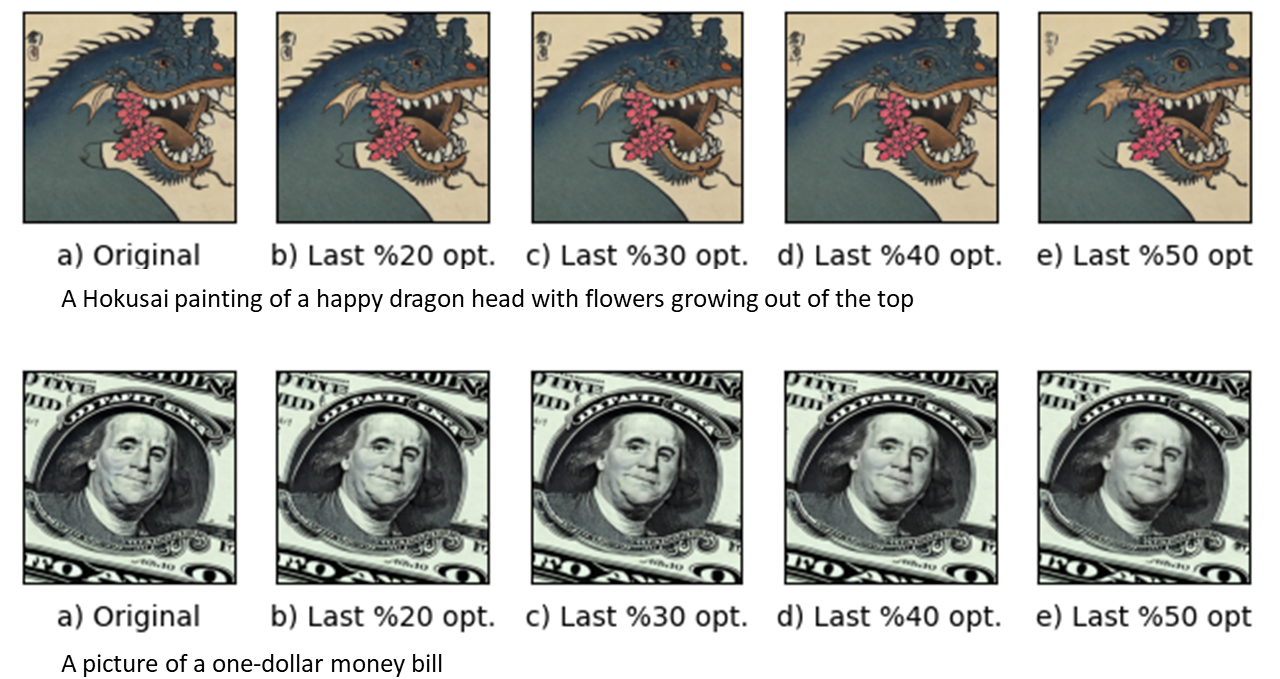}
  \caption{Image \textit{a} is the baseline with no optimizations and \textit{b}, \textit{c}, \textit{d} and \textit{e} are the same image generated applying optimization to the last 20\%, 30\%, 40\%, and 50\% of the iterations.}
  \label{SBS-opt-range}
\end{figure}

Figure \ref{SBS-opt-range} shows a side-by-side (SBS) comparison of five images generated for each prompt. 
When comparing images in the same row, we can observe a degradation in quality as the degree of optimization increases from left to right. Upon examining images \textit{a} and \textit{b}, we found that optimizing the last 20\% of the iterations has almost no impact on image quality. Moreover, images \textit{a} and \textit{e} demonstrate that we can extend optimization to the last 50\% of the iterations and still maintain a visually appealing result. Based on these observations, we can conclude that the threshold for noticeable changes in image quality lies at approximately 20\% optimization (as demonstrated by image \textit{b}). We further investigated this claim in the subsequent subsection.

\subsection{Threshold for optimization}
We developed a SBS test comprising 60 prompts listed in~\tref{table-prompts}. For each prompt, we generated two images: one without optimization and the other with 20\% of iterations optimized. We then enlisted six participants to evaluate the image pairs and select the one they deemed superior in terms of quality. Participants were blinded to which image had been optimized and asked to make their decision solely based on their personal perception of image quality. 
\begin{figure}
  \centering
  \includegraphics[width=0.4\textwidth]{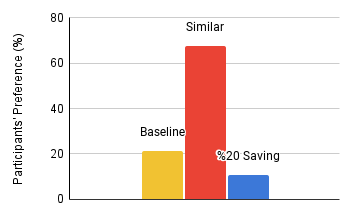}
  \caption{In our SBS test, 68\% think the optimized and non optimized images are similar and 21\% prefer the non-optimized while the remaining 11\% prefer the optimized image.}
  \label{SBS-results}
\end{figure}
The SBS results depicted in Figure \ref{SBS-results} indicate that 68\% of participants perceived the two images as being similar. However, only 21\% of participants preferred the Baseline image, while the remaining 11\% preferred the optimized version.
\subsection{Performance gain}
To evaluate performance gains, we generated the same image multiple times using the same prompt. The first 10 image generations were conducted for warming up purposes, after which we measured the time required to generate 50 images using different seeds and calculated the average time required. The results of these measurements are summarized in Table \ref{table-perf}. 
\begin{table}[!h]
\begin{tabular}{|c|c|c|c|c|c|}
\hline
Iterations optimized & No opt. & 20\% of iters & 30\% of iters & 40\% of iters & 50\% of iters \\
\hline
Time(s) & 9.94 & 9.13 & 8.74 & 8.33& 7.92 \\
\hline
Saving & - & 8.2\% & 12.1\% & 16.2\% & 20.3\% \\

\hline
\end{tabular} 

\caption{The average measured time to generate an image using SD inference pipeline on Tesla V100 running CUDA version 11.6 and denoising iteration set to 50.}
\label{table-perf}
\end{table}

As anticipated, the speed-up observed was approximately half of the number of iterations that had been optimized. This is because the denoising Unet comprises the bulk of the computation, and by optimizing a given number of iterations, we were effectively cutting the Unet computation time in half for those iterations. 

\subsection{Guidance scale tuning}
\begin{figure}
  \centering
  \includegraphics[width=1.0\textwidth]{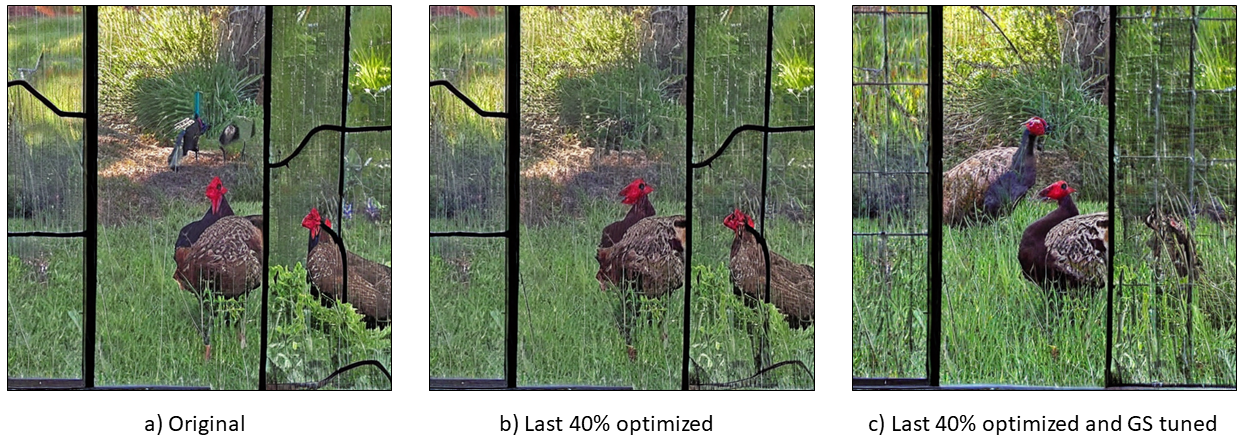}
  \caption{Images generated for the same prompt: "Wild turkeys in a garden seen from inside the house through a screen door". The three images show case loosing some details applying aggressive optimization then regaining lost details through tuning the GS.}
  \label{GS-tuning}
\end{figure}

To enhance the performance gain, we recommend tuning the guidance scale (GS) after implementing optimization on a significant number of iterations. An illustration of this approach is presented in Figure \ref{GS-tuning}. A comparison between results \textit{a} and \textit{b} demonstrates that applying optimization to a large portion of the iterations (40\% here) can lead to the loss of important details, such as the third bird in the distance. However, by adjusting the GS (from 7.5 to 9.6 in this case), we were able to restore the lost details, including the third bird. 
We acknowledge that due to time constraints, we are only able to provide a demonstration of this technique and encourage readers to explore this opportunity further in order to improve image quality.
\section{Conclusion}
\label{sec:conclusions}
In this study, we aimed to optimize the SD inference pipeline, which involves a denoising loop that iterates between 50 to 200 times for the SD model. We conducted experiments to evaluate the effect of optimization on this denoising loop and found that: 1) later iterations of the denoising loop are less sensitive to optimization, making it beneficial to apply optimizations to these iterations while keeping the earlier iterations unchanged. 2) We can optimize the later iterations by limiting the noise computation in the denoising loop to conditional noise, cutting the denoising loop computation time in half for those iterations by eliminating the computation of unconditional noise. Our experiments revealed that optimizing 20\% of the iterations resulted in an image quality similar to that of the unoptimized image, while saving approximately 8.2\% of the inference time. We further extended the optimization to 50\% of the iterations, which resulted in a 20.3\% reduction in inference time while still generating images of good quality for human perception.


{
\bibliographystyle{plain}
\bibliography{ref.bib}
}

\clearpage
\onecolumn
\appendix


\begin{longtable}{|c|p{15cm}|}
\hline
1 & An armchair in the shape of an avocado \\
2 & An old man is talking to his parents \\
3 & A grocery store refrigerator has pint cartons of milk on the top shelf, quart cartons on the middle shelf, and gallon plastic jugs on the bottom shelf \\
4 & An oil painting of a couple in formal evening wear going home get caught in a heavy downpour with no umbrellas \\
5  & Paying for a quarter-sized pizza with a pizza-sized quarter \\
6  & Wild turkeys in a garden seen from inside the house through a screen door \\
7  & A watercolor of a silver dragon head \\
8  & A watercolor of a silver dragon head with flowers \\
9  & A watercolor of a silver dragon head with colorful flowers \\
10  & A watercolor of a silver dragon head with colorful flowers growing out of the top \\
11  & A watercolor of a silver dragon head with colorful flowers growing out of the top on a colorful smooth gradient background \\
12  & A red basketball with flowers on it, in front of blue one with a similar pattern \\
13  & A Cubism painting of a happy dragon with colorful flowers growing out of its head \\
14  & A cyberpunk style illustration of a dragon head with flowers growing out of the top with a rainbow in the background, digital art \\
15  & A Hokusai painting of a happy dragon head with flowers growing out of the top \\
16  & A Salvador Dali painting of 3 dragon heads \\
17  & A Leonardo Da Vinci painting of 3 dragon heads and 2 roses \\
18  & 3d rendering of 5 tennis balls on top of a cake \\
19  & A person holding a drink of soda \\
20  & A person is squeezing a lemon \\
21  & A person holding a cat \\
22  & A red ball on top of a blue pyramid with the pyramid behind a car that is above a toaster \\
23  & A boy is watching TV \\
24  & A photo of a person dancing in the rain \\
25  & A photo of a boy jumping over a fence \\
26  & A photo of a boy is kicking a ball \\
27  & A path in a forest with tall trees \\
28  & A sunset with a cloudy sky and a field of grass \\
29  & A dirt road that has some grass on it \\
30  & A beach with a lot of waves on it \\
31  & A road that is going down a hill \\
32  & A rocky shore with waves crashing on it \\
33  & Abraham Lincoln touches his toes while George Washington does chin-ups  Lincoln is barefoot \\
34  & A snowy forest with trees covered in snow \\
35  & A path in a forest with tall trees \\
36  & A path through a forest with fog and trees \\
37  & A field with a lot of grass and mountains in the background \\
38  & A waterfall with a tree in the middle of it \\
39  & A foggy sunrise over a valley with trees and hills \\
40  & A beach with a cloudy sky above it \\
41  & A black and white photo of a mountain range \\
42  & A mountain range with snow on top of it \\
43  & A picture of a one-dollar money bill \\
44  & Supreme Court Justices play a baseball game with the FBI \\
45  & A picture of a Red Robin \\
46  & A picture of Coco Cola can \\
47  & A picture of Costco store \\
48  & A high-quality photo of a golden retriever flying a yellow floatplane \\
49  & A profile photo for a smart, engaging digital assistant \\
50  & A picture of a multilingual Bert hanging out with Elmo and Ernie \\
51  & A molecular diagram showing why ice is less dense than water \\
52  & A historical painting showing the invention of the wheel \\
53  & A picture of water pouring out of a jar in outer space \\
54  & Futuristic view of Delhi when India becomes a developed country as digital art \\
55  & A donkey and an octopus are playing a game  The donkey is holding a rope on one end, the octopus is holding onto the other  The donkey holds the rope in its mouth \\
56  & A mirrored view of the Great Sphinx of Giza as digital art \\
57  & Concept art of the next generation cloud-based game console \\
58  & A silver dragon head \\
59  & A pear cut into seven pieces arranged in a ring \\
60  & A tomato has been put on top of a pumpkin on a kitchen stool. There is a fork sticking into the pumpkin \\
61  & An elephant is behind a tree \\

\hline
\caption{Prompts utilized to generate images for the SBS test.}
\label{table-prompts}
\end{longtable}

\end{document}